\documentclass[letterpaper, 10 pt, journal, twoside]{IEEEtran}

\IEEEoverridecommandlockouts                              

\usepackage{graphicx} 
\usepackage{amsmath} 
\usepackage{amssymb}  
\usepackage{booktabs}
\usepackage[dvipsnames]{xcolor}
\usepackage{paralist}
\usepackage{hyperref}

\usepackage{cleveref}
\usepackage{multirow}
\usepackage{balance}
\usepackage{subfigure}
\usepackage[noadjust]{cite}
\usepackage{adjustbox}
\usepackage{paralist}

\let\labelindent\relax
\usepackage{enumitem} 

\newcommand{\myparagraph}[1]{\vspace{2.0pt}\noindent\textbf{#1.}}

\title{
JIST: Joint Image and Sequence Training\\for Sequential Visual Place Recognition
}
\author{Gabriele Berton, Gabriele Trivigno, Barbara Caputo and Carlo Masone
\thanks{Manuscript received 13 July 2023; accepted 28 November 2023. Date of publication; date of current version. This letter was recommended for publication by Associate Editor Y. Liao and Editor S. Behnke upon evaluation of the reviewers’ comments. This work was supported by Consorzio Interuniversitario Nazionale per l’Informatica (CINI), Roma, Italy. Gabriele Berton and Gabriele Trivigno contributed equally to this work. Gabriele Berton is the corresponding author.}
\thanks{
The authors are with the Visual and Multimodal Applied Learning Lab, Department of Control and Computer Engineering, Politecnico di Torino, 10138 Torino, Italy (e-mail: gabriele.berton@polito.it; gabriele.trivigno@polito.it; barbara.caputo@polito.it; carlo.masone@polito.it).}%
\thanks{The code is available at \url{https://github.com/ga1i13o/JIST}}
\thanks{Digital Object Identifier 10.1109/LRA.2023.3339058}
}

\begin{document}

\maketitle

\begin{abstract}

Visual Place Recognition aims at recognizing previously visited places by relying on visual clues, and it is used in robotics applications for SLAM and localization.
Since typically a mobile robot has access to a continuous stream of frames, this task is naturally cast as a sequence-to-sequence localization problem.
Nevertheless, obtaining sequences of labelled data is much more expensive than collecting isolated images, which can be done in an automated way with little supervision. As a mitigation to this problem, we propose a novel Joint Image and Sequence Training protocol (JIST) that leverages large uncurated sets of images through a multi-task learning framework.
With JIST we also introduce SeqGeM, an aggregation layer that revisits the popular GeM pooling to produce a single robust and compact embedding from a sequence of single-frame embeddings.
We show that our model is able to outperform previous state of the art while being faster, using 8 times smaller descriptors, having a lighter architecture and allowing to process sequences of various lengths.

\end{abstract}



\section{Introduction}
Localization is a fundamental functionality for autonomous mobile robots, and one of its key ingredients is Visual Place Recognition (\textbf{VPR})~\cite{Masone_2021_survey}, i.e., the task of matching a current visual observation (an image or video stream) to previously visited places. For example, VPR is used for loop closure detection in SLAM~\cite{Konstantinos-2022_loopclosure}, for re-localization in the kidnapped robot problem \cite{Angeli-2008-relocalization} and also for pure localization when a map is already available \cite{Pion_2020_benchmark} and when GNSS measurements are  precluded \cite{Yudin_2023_indoor_vpr, Zeng_2017_underground_vpr}.
Additionally, VPR is used to select rough candidates for precise 6-DoF pose estimation (i.e., visual localization)~\cite{Pion_2020_benchmark,Toft_2020_visuallocalization_net}.

Across these robotics applications, VPR is typically performed using methods that process short sequences of images acquired by cameras onboard the robot - what is called \textbf{sequence-to-sequence} or \textbf{seq2seq} place recognition \cite{Warburg_2020_msls}.
A recent trend in this sense is to frame the seq2seq problem as a retrieval task on learnt embeddings (\emph{sequence descriptors}) that represent entire sequences rather than individual frames \cite{Facil_2019_multiViewNet,Garg_2020_deltaDescr,Garg_2021_seqNet,Mereu_2022_seqvlad}. 
This new paradigm not only intrinsically captures the temporal information in the video stream, but it is also more efficient than individually matching each frame with previous observations~\cite{Garg_2021_seqNet,Mereu_2022_seqvlad}.
However, the accuracy and robustness achieved by sequence descriptors is bounded by the limited availability of large datasets of sequences. 
Indeed, for the classic \textbf{image-to-image} VPR (\textbf{im2im}~\cite{Warburg_2020_msls}) the availability of massive datasets has been instrumental in setting the latest state of the art \cite{Berton_2022_cosPlace, Alibey_2023_mixvpr}, producing descriptors that generalize better and are very compact\footnote{This also entails a reduced latency, because the size of descriptors has a linear correlation with the matching time of the retrieval~\cite{Berton_2022_benchmark}.}. 
Yet, due to difficulties in curating sequences~\cite{Warburg_2020_msls, Philbin_2007_oxford5k}, the largest dataset currently available for the seq2seq task (Mapillary Street Level Sequences \cite{Warburg_2020_msls}) is 40$\times$ smaller than the largest datasets for image-to-image VPR~\cite{Berton_2022_cosPlace, Martinez_2020_Pit30M}.

\begin{figure}[t!]
    \begin{center}
    \begin{minipage}{.30\textwidth}
        \includegraphics[width=\textwidth]{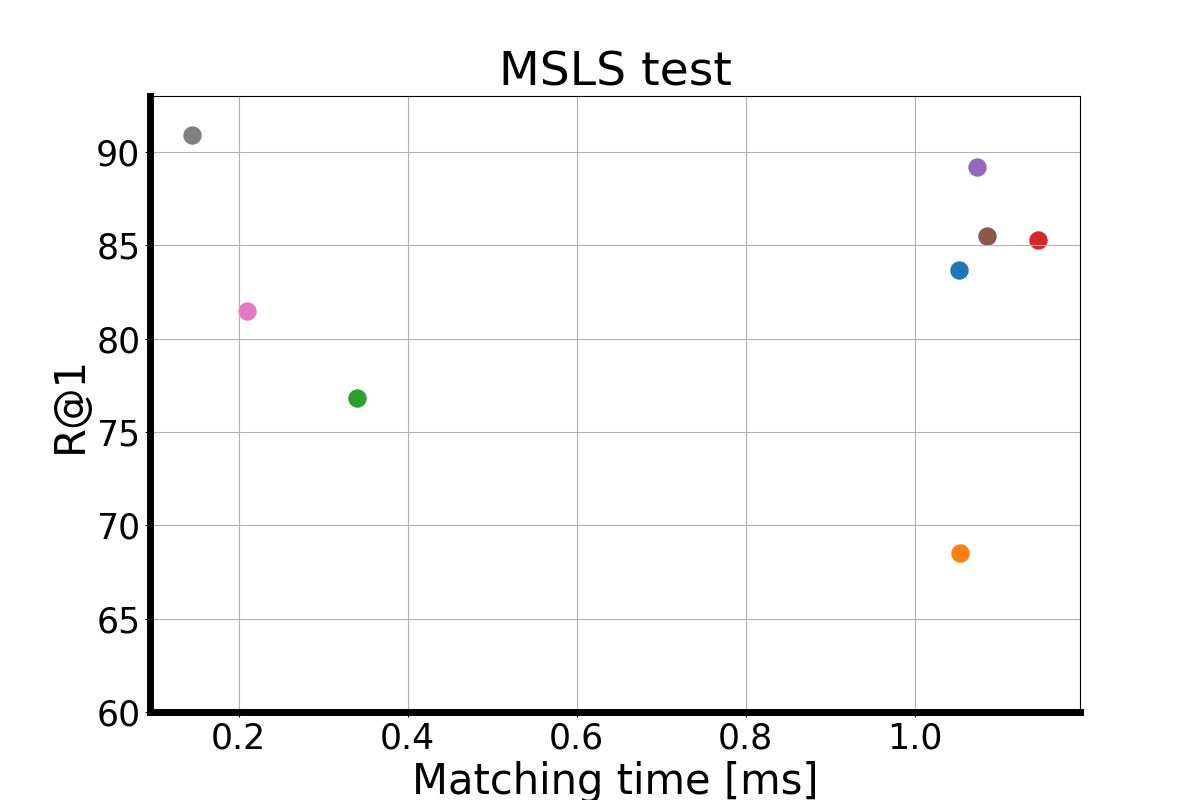}
    \end{minipage}
    \begin{minipage}{.17\textwidth}
        \includegraphics[width=\textwidth]{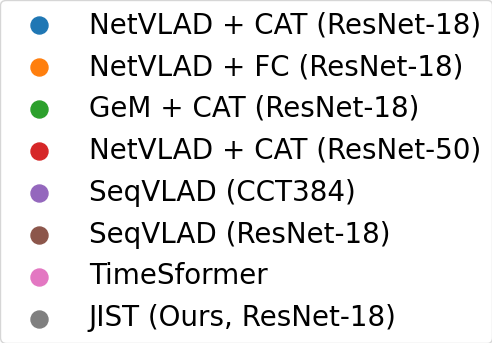}
    \end{minipage}
    \end{center}
\caption{Our multi-task training framework allows to surpass previous SOTA in performance. Thanks to our novel layer SeqGeM, we are able to cut down the matching time by an order of magnitude.}
\label{fig:teaser}
\end{figure}
Given the correlation between the seq2seq and the im2im tasks, we argue that it is possible to produce more effective sequence descriptors by jointly training a model not only on sequences, but also on the readily available massive datasets for image-to-image VPR:
on one hand, the im2im training from huge-scale datasets would improve the model's generalizability; on the other, sequence-to-sequence learning would embed the model with robustness to sequentially changing scenes and teach it how to temporally aggregate frame-level information.
To this end, we propose a new training methodology that jointly uses images and sequences and exploits a state-of-the-art architecture originally developed for im2im VPR to first extract discriminative embeddings from individual frames and then aggregate them.
While this new training method enables the model to effectively learn also from large datasets for the im2im tasks, it does not automatically solve the issue of large-dimensional embeddings required by previous SOTA \cite{Mereu_2022_seqvlad}.
To address this issue, in \cref{sec:seq_gem} we introduce a new aggregation layer called \textbf{SeqGeM}, that revisits the popular generalized mean pooling~\cite{Radenovic_2019_gem} by applying it along the temporal axis, resulting in very compact descriptors and, consequently, speeding-up the matching time
(see \cref{fig:teaser}).
The combination of this training method and SeqGeM takes the name of Joint Image and Sequence Training, or \textbf{JIST}.

To summarize, we bring the following contributions:
\begin{itemize}
    \item 
    We propose a novel multi-task training framework to leverage existing large scale datasets of image-to-image VPR and improve upon the seq2seq task~\cite{Warburg_2020_msls};
    \item We introduce the SeqGeM aggregation layer, which revisits the popular generalized mean pooling~\cite{Radenovic_2019_gem} by aggregating individual frames descriptors along the temporal axis and resulting in compact and robust descriptors regardless of the input sequence length;
    \item We show that, compared to previous SOTA, our pipeline achieves better results and faster inference thanks to its reliance on smaller dimensional descriptors.
\end{itemize}

\section{RELATED WORKS}

\myparagraph{Sequence matching}
Sequence matching, or frame-by-frame matching, represents an established approach to seq2seq \cite{Ho_2007_first_seq_search,Milford_2012_seqslam}, and it operates by building a similarity matrix wherein descriptors of single query frames are compared to database ones. The best match is then determined by aggregating the scores in the matrix under simplifying assumptions, such as constant velocity  or no stops \cite{Schubert_2021_vpr_hard}, which makes it hard to generalize to real-world applications. 
There is a rich literature on sequence matching that tries to relax these assumptions by exploiting ego-motion information or using complex methods \cite{Naseer_2018_vl_across_seasons} and graph-based frameworks~\cite{Vysotska-2016_lazymatching,Schubert-2021_graphbased,schubert-2021_fast}.
Recently, SeqMatchNet \cite{Garg_2021_SeqMatchNetCL} has also addressed  the fact that these methods rely on learned image-to-image descriptors trained without considering the downstream procedure of score aggregation.
Despite these improvements, sequence matching can generally be expensive to perform, as it requires each frame from the query to be matched to each frame of all databases sequences, as discussed in~\cite{Garg_2021_seqNet, Mereu_2022_seqvlad}.

\myparagraph{Sequence descriptors}
Sequence descriptor methods summarize each sequence with a single embedding which can be used for retrieving the most similar matches.
This allows to incorporate temporal clues directly into the descriptors and to perform the similarity search directly on sequences rather than frames, thus greatly reducing the matching time. 
Facil et al. \cite{Facil_2019_multiViewNet} first introduced the idea of sequence descriptors in VPR using simple aggregation techniques such as concatenation, sum, or processing via a LSTM network. \cite{Warburg_2020_msls} extended their benchmark on the Mapillary Street Level Sequences (MSLS) dataset.
A non-learnable aggregation via discrete convolution was explored by Garg et al.~\cite{Garg_2020_deltaDescr}.
Alternatively, 1D temporal convolutions were employed in SeqNet \cite{Garg_2021_seqNet} to obtain a learnable aggregation of frame descriptors. Recently, \cite{Neubert-2021_hyperd} demonstrated a hyperdimensional computing approach
to systematically combine information from multiple single-image descriptors.
Considering the architectural differences among these methods, 
\cite{Mereu_2022_seqvlad} provides a benchmark and taxonomy for seq2seq methods depending on how the frame-level features are fused together, and then it introduces the SeqVLAD aggregation layer that achieves SOTA performance.
A follow up work is found in \cite{zhang_2023_spatiotemp}.

\myparagraph{Image-to-image place recognition on large databases}
There is a parallel body of literature in computer vision on image-to-image place recognition, 
addressing it as a retrieval task using global image descriptors.
For years, the de-facto standard method has been NetVLAD \cite{Arandjelovic_2018_netvlad}, that also introduced the training procedure with mining and triplet loss. 
However, recently \cite{Berton_2022_benchmark} has pointed out that the cost of mining triplets
is a major bottleneck that prevents these methods from scaling to large datasets.
This consideration inspired few recent papers to pursue mining-free methods in order to enable training on massive datasets.
Firstly, CosPlace \cite{Berton_2022_cosPlace} provides a method to split large dense datasets into non-overlapping classes, which then allows for training to be performed with scalable loss functions.
Using a different approach, Ali-Bey et al. \cite{Alibey_2022_gsvcities} provides a dataset that is already split into well-defined classes, allowing to use standard retrieval losses without the need for mining.
MixVPR \cite{Alibey_2023_mixvpr} uses a similar training approach, and shows that well-designed architectures can provide a boost in recall.
Most recently, \cite{Leyvavallina_2021_gcl} introduces a novel reward function, named Generalized Contrastive Loss, to dispense from hard-pair mining.

This trend in the literature shows that a method that is able to efficiently leverage large scale datasets can bring great benefits for performances. 
In seq2seq VPR this has not been possible because it is hard to obtain such large datasets. 
In this paper we propose a training protocol for sequence descriptors that is able to leverage the large amount of data readily available for the im2im task, even though it does not contain sequences of frames.
Moreover, we show that our approach is able to improve upon previous SOTA while reducing the cost of deployment.

\section{METHOD}

\begin{figure*}[t!]
  \centering
  \includegraphics[width=\textwidth]{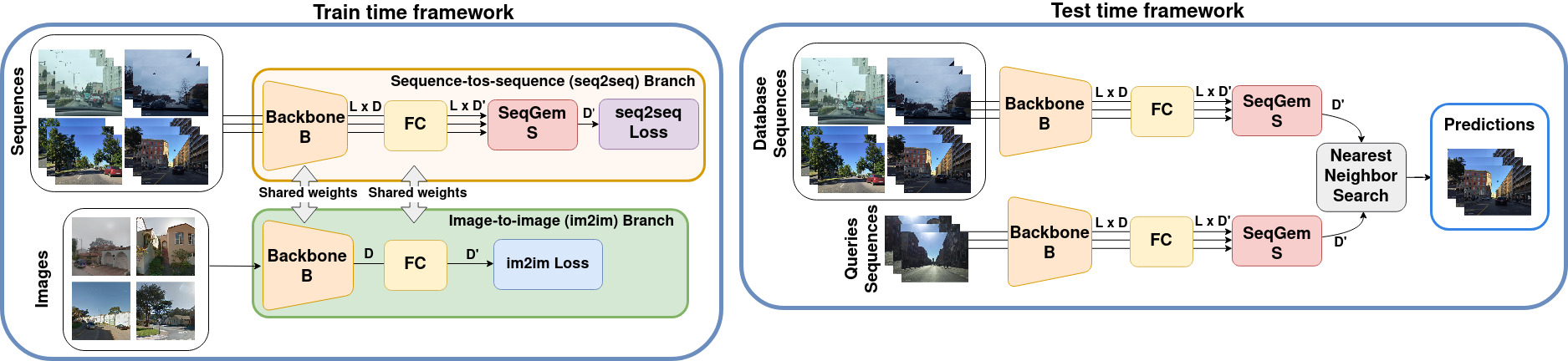}
  \vspace{-0.1cm}
  \caption{\textbf{Overview of the JIST framework.}
  At training time (left) we use two branches, one for sequences and one for single-images. Each branch has a separate loss, while sharing part of their weights.
  The multi-task training allows to obtain discriminative frame-wise embeddings by exploiting the powerful representations learned by the backbone and fully connected  from single images.
  At test time (right) we only use the sequences branch, and we follow the standard image retrieval pipeline: embeddings are extracted for both database and queries sequences, and then a prediction for database sequence that is most similar to the query is computed through a kNN.
  Note that in a real-world scenario, the potentially expensive embeddings extraction for database sequences can be performed offline, making the framework fast (more information on efficiency in \cref{sec:realworld_deployment}).
  }
  \label{fig:architecture}
  \vspace{-2mm}
\end{figure*}

\subsection{Problem setting}
We tackle the task of \emph{seq2seq VPR} that is formally defined in~\cite{Warburg_2020_msls}: given a query sequence the system has to output a sequence from the available database that matches the former.
Since the database sequences have GPS labels, this allows to infer an estimate of the query's position.
A match is deemed correct if any of the retrieved frames is within 25 meters~\cite{Warburg_2020_msls} from any of the query frames. The common recall@N metric \cite{Berton_2022_benchmark, Garg_2021_seqNet, Milford_2012_seqslam, Mereu_2022_seqvlad} is used as an aggregate evaluation, and it represents the percentage of queries that have at least one correct match in the top-N candidates.

Our method builds on the idea that the task of seq2seq VPR can be split into 2 learning objectives: (i) learn to extract  features that are distinctive for localization (\emph{i.e.} ignore transient objects, focus on static components, their style and relative position) and (ii) model the temporal evolution of these salient features within a sequence.
In the spirit of deep learning, it is possible to jointly acquire both capabilities in an end-to-end fashion from a dataset of sequences  \cite{Mereu_2022_seqvlad}. 
However, we observe that for the first objective we do not necessarily need sequences, but we can exploit existing large-scale non-sequential VPR datasets to embed into our model robustness to a large variety of scenarios. Following this intuition, we devise our multi-task learning framework.

\subsection{Multi Task Framework: Overview}
The typical descriptor extractor architecture for im2im retrieval is composed by a backbone and an aggregator of feature maps (or tokens). For retrieval on sequences, there is an additional step to aggregate frame-level information \cite{Mereu_2022_seqvlad}.
In order to leverage both im2im and seq2seq datasets, we need a unified architecture with a frame-level aggregation layer able to process both individual images and sequences.
Thus, we propose a novel double-branched architecture: one branch takes sequential data as input, while the other takes single images (see \cref{fig:architecture}).
We iteratively feed each branch with one batch of its corresponding input, compute their respective losses, backpropagate through the entire model and sum the gradients computed for each loss, which are then used for optimization.
In doing so, we ensure that both branches share the same gradients and weights: in practice this makes the backbone and fully connected layer (FC) of the two branches identical, and allows joint optimization on both losses in a Siamese-like fashion.
Following, we explain how the two branches work at inference and training time.


\subsection{Sequence-to-sequence branch}

The sequence-to-sequence branch has the objective of 
exploiting all the frame-wise information extracted by the im2im branch, via the shared backbone and FC layer, while learning to aggregate temporal information from sequences into compact descriptors.
The input to this branch is formed by sequences $x_{seq}$ of frames, with $x_{seq} \in \mathbb{R}^{L \times H \times W \times C}$ (where $L$ is the sequence length).
The sequences are passed to a backbone $B$, which extracts $L \times D$ dimensional features
where the $D$ depends on the backbone.
These features are then passed through an FC layer $F$, which acts as a whitening transformation \cite{Radenovic_2019_gem}
and produces $L$ \; $D'$-dimensional descriptors (\emph{i.e.} one descriptor per frame), where $D'$ can be set to a chosen output dimension.
At this point, we need two more ingredients: firstly a \emph{sequence aggregation} module that combines these frames into a single vector (the sequence descriptor); secondly, a \emph{loss function} for the seq2seq task. 

\myparagraph{Sequence aggregation: SeqGeM}
\label{sec:seq_gem}
The current SOTA sequence descriptor from~\cite{Mereu_2022_seqvlad} is built using the SeqVLAD aggregator. This module reinterprets the classic NetVLAD module~\cite{Arandjelovic_2018_netvlad} to make it suitable for sequences. In a nutshell, given a set of $D'$-dimensional input descriptors SeqVLAD produces a single sequence descriptor vector of size $K\cdot D$, where $K$ is a parameter indicating the number of clusters used to summarize the input vectors.
In practice, the implementation from~\cite{Mereu_2022_seqvlad} uses $K = 64$, which significantly increases the size of the sequence descriptor and, as a consequence, the matching time of the retrieval. To mitigate this problem, \cite{Mereu_2022_seqvlad} uses a PCA compression operation, which nevertheless adds a post-processing computational overhead.

\begin{figure}
  \centering
  \includegraphics[width=\linewidth]{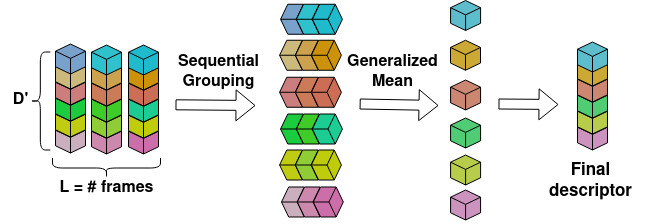}
  \vspace{-0.1cm}
  \caption{Sketch of our proposed SeqGeM layer. Given D-dimensional feature vectors from L frames, SeqGeM produces a single descriptor/embedding of dimensionality D, which contains information from the whole sequence. 
  }
  \label{fig:seqgem}
  \vspace{-0.2cm}
\end{figure}

For all these reasons, we propose a new aggregation module called Sequential Generalized Mean (\textbf{SeqGeM}), which revisits the popular GeM layer~\cite{Radenovic_2019_gem} to operate on sequences, by applying its pooling operation along the temporal axis given a sequence of single-image embeddings (see \cref{fig:seqgem}). 
Formally, the SeqGeM layer is defined as 
\begin{equation}
    \begin{aligned}
        S: \mathbb{R}^{L \times D'} & \rightarrow \mathbb{R}^{D'} \\
           [d_0, \ldots d_{L-1}] & \mapsto \left( \frac{1}{L} \sum_{i = 0}^{L-1} d_i ^p \right) ^{\frac{1}{p}}
    \end{aligned}
\end{equation}
where $p$ is a learnable parameter and $d_i$ is the descriptor of the $i^{th}$ frame.
Therefore, the sequence descriptor extraction process is
\begin{equation}
    f_{seq} = S(F(B(x_{seq})))
\end{equation}

SeqGeM is implemented with differentiable operations, and it has a few desirable properties: i) it natively produces low-dimensional descriptors without requiring a PCA compression; ii) it is learnable; iii) it has few parameters; iv) it is flexible w.r.t. the length of input sequences, so that sequences of different length can be compared to each other.
Finally, SeqGeM is purposefully designed to aggregate only the final descriptors of each frame, instead of the frame's feature maps, as in this way it is able to (i) take advantage of the entire im2im branch, which is trained on large amount of images, and (ii) take as input small descriptors and produce small outputs, whereas usually methods that take as inputs the feature maps (e.g. SeqVLAD) produce large-dimensional sequence descriptors which increases memory and time requirements.

\myparagraph{Seq2seq loss}
Following best practices from the literature, we use the popular weakly supervised margin triplet loss \cite{Arandjelovic_2018_netvlad}, which takes a query, its positive (a sequence from the same place), and a negative.
For best results, negatives need to be \textit{mined}, because selecting random negatives would lead to trivial triplets (i.e., with loss 0), by selecting the negatives closest to the queries in features space.
Given triplets of query, positive and negative, the weakly supervised triplet loss, used to train the seq2seq branch, is defined as:
\begin{equation}
    \mathcal{L}_{seq2seq} = \sum_i max(0, d(f_{seq}^{q_i}, f_{seq}^{p_i}) - d(f_{seq}^{q_i}, f_{seq}^{n_i}) + m))
\end{equation}
where $f_{seq}^{q}, f_{seq}^{p}, f_{seq}^{n}$ represent the features of a query, its positive and negative, $m$ is the margin of the triplet loss, and $d(\cdot)$ is the euclidean distance between two features.

\subsection{Image-to-image branch}
The second branch processes single images instead of sequences: 
given input images $x_{im} \in \mathbb{R}^{H \times W \times C}$ the image branch produces $D'$ dimensional local feature descriptors which can be fed to the image loss.
The local feature descriptors are computed as
\begin{equation}
    f_{im} = F(B(x_{im}))
\end{equation}
where the backbone $B$ and fully connected layer $F$ are shared with sequence-to-sequence branch (see \cref{fig:architecture}).
Finally, we attach a loss $\mathcal{L}_{im2im}$ for the image-to-image  task, that backpropagates through $B$ and $F$.

\myparagraph{Im2im loss}
Since our goal for this branch is to exploit huge datasets of single images to learn robust representations, we resort to the CosPlace training protocol and loss~\cite{Berton_2022_cosPlace} that is the current state-of-the-art for large scale im2im VPR and was designed to be used on the massive San Francisco eXtra Large (SF-XL) dataset.
Below we provide a summarized explanation of the CosPlace training protocol, although we note that this is not meant to be a thorough description and we refer the reader to the original CosPlace paper~\cite{Berton_2022_cosPlace} for a more detailed explanation.

The CosPlace training protocol is divided in two steps. In the first step, the SF-XL dataset which contains images labeled with UTM coordinates and heading angles is partitioned into classes based on their position and orientation. This process, that is performed once prior to the actual training, divides the geographical area into small squared cells ($10 \times 10$ meters)
and splits each cell into 12 classes along the orientation/heading (\emph{i.e.} each class is  $30^\circ$ wide), thus ensuring that all the images in a single class view the same scene (by having similar position and orientation).
This division of the continuous label space in a finite number of classes enables the usage of highly scalable losses for large-scale image retrieval, such as the CosFace loss~\cite{Wang_2018_cosFace}.

Therefore, the second step consists in training the model using the CosFace loss on the obtained classes. 
However, naively using all the classes would be problematic, because images in two adjacent classes may have a very high visual overlap, thus potentially containing the same scene seen from slightly different points of view. Since this would lead to unstable gradients during optimization, the training protocol only considers images from a subset of classes chosen so that no two adjacent classes are used at the same time. This subset is not fixed, but it is changed iteratively during training, to allow the model to see all the images in the dataset.

Summarizing, in this paper we denote as $\mathcal{L}_{im2im}$ the CosFace loss applied according to the CosPlace protocol.
However, we want to remark that in principle our multi-task framework is loss-agnostic, so the im2im loss can be easily swapped with another one, for example if a more performing loss becomes available.

\subsection{Total multi-task loss}
Overall, the total loss of our multi-task framework is 
\begin{equation}
    \mathcal{L}_{multi-task} = \lambda_{seq2seq} \mathcal{L}_{seq2seq} + \lambda_{im2im} \mathcal{L}_{im2im}
\end{equation}
where $\lambda_{seq2seq}$ and $\lambda_{im2im}$ are hyperparameters.
The combination of this multi-task loss with architecture that includes the SeqGeM aggregation makes our multi-task framework, which we name Joint Image and Sequence Training, or JIST.

\section{Experiments}

\subsection{Experimental setup}

\begin{table*}[th!]
\caption{
Evaluation of sequential descriptors and sequence matching, on sequences of length 5: recall@1 on various datasets.
SL stands for Sequence Length, CAT indicates concatenation of descriptors, FC stands for Fully Connected layer.
Extraction time is the time to extract descriptors/embeddings, and matching time is the time to find the predictions given the descriptors given the test database of MSLS (with 13584 sequences).
Both times refer to a single query.
{*} denotes a non-trained method. Best results in bold, second best are underlined.}
\vspace{-0.1cm}
\label{tab:main_tab}
\begin{center}
\resizebox{\textwidth}{!}{
\begin{tabular}{lccccccccc}
\toprule
\multirow{2}{*}{\begin{tabular}[c]{@{}c@{}}Method\end{tabular}} &
\multirow{2}{*}{\begin{tabular}[c]{@{}c@{}}Backbone\end{tabular}} &
\multirow{2}{*}{\begin{tabular}[c]{@{}c@{}}Descriptor\\ Dimension\end{tabular}} &
\multirow{2}{*}{\begin{tabular}[c]{@{}c@{}}GPU Memory\\Occupation (GB)\end{tabular}}&
\multirow{2}{*}{\begin{tabular}[c]{@{}c@{}}Extraction\\Time (ms)\end{tabular}}&
\multirow{2}{*}{\begin{tabular}[c]{@{}c@{}}Matching\\Time (ms)\end{tabular}}&
\multirow{2}{*}{\begin{tabular}[c]{@{}c@{}}Train on Melbourne\\ Test on MSLS\end{tabular}}&
\multirow{2}{*}{\begin{tabular}[c]{@{}c@{}}Train on MSLS\\ Test on MSLS\end{tabular}} &
\multirow{2}{*}{\begin{tabular}[c]{@{}c@{}}Train on RobotCar\\Test on RobotCar\end{tabular}} & 
\multirow{2}{*}{\begin{tabular}[c]{@{}c@{}}Train on MSLS\\Test on RobotCar\end{tabular}}
\\
&&&&&&&&\\
\midrule
SeqSLAM{*} \cite{Milford_2012_seqslam}  & VGG-16    & 4096 $\cdot$ SL& 2.68  &  39.8 & 500.1 & 45.9 &  45.9 & 34.7 & 34.7 \\
HVPR \cite{Garg_2021_seqNet}             & VGG-16    & 4096 $\cdot$ SL & 2.68  & 39.8 & 0.9  & 51.0 & - & 56.8 & - \\
SeqMatchNet\cite{Garg_2021_SeqMatchNetCL}&VGG-16     & 4096 $\cdot$ SL & 2.68  & 39.8 & 500.1 & 44.8 & - & 51.9 & - \\
Delta Descriptors{*}\cite{Garg_2020_deltaDescr} & VGG-16    & 4096 $\cdot$ SL & 2.68 & 39.8 & 1.1 &43.0& 43.0 & 18.0 & 18.0 \\
GeM + CAT \cite{Warburg_2020_msls}     & ResNet-18 & 256 $\cdot$ SL     & 2.04 & 10.6 & 0.3 & 66.7 & 76.8 & 75.4 & 26.4 \\
GeM + CAT \cite{Warburg_2020_msls}     & ResNet-50 & 1024 $\cdot$ SL     & 2.25 & 29.5 & 1.8 & 63.4 & 68.6 &81.3 & 14.1 \\
NetVLAD + CAT + PCA \cite{Facil_2019_multiViewNet}                  & ResNet-18 & 4096     & 2.04  & 10.4 & 1.1 & 75.5 & 83.7 & 67.8 & 47.0 \\
NetVLAD + CAT + PCA \cite{Facil_2019_multiViewNet}                  & ResNet-50 &4096       & 2.26  & 30.4 & 1.1 & 74.9 & 85.3 & 89.3 & 62.4 \\
SeqPool + CAT \cite{Hassani_2021_cct}                       & CCT384    &384 $\cdot$ SL& 2.01  & 15.6 & 0.5 & 69.9 & 77.8 & 77.4 & 42.6 \\
SeqNet \cite{Garg_2021_seqNet}          & VGG-16    & 4096     & 2.68  & 39.8 & 1.1 & 50.1 & - & 60.5 & - \\
SeqNet \cite{Garg_2021_seqNet}         & ResNet-50 & 4096     & 2.26  & 31.0 & 1.1& 45.6 & - & 61.3 & - \\
NetVLAD + FC \cite{Facil_2019_multiViewNet}                          & ResNet-18 & 4096     & 3.39  & 10.5 & 1.1 & 55.5 & 68.5 & 44.7 & 19.3 \\
SeqVLAD + PCA \cite{Mereu_2022_seqvlad}                          & ResNet-18 & 4096     & 2.04  & 10.5 & 1.1 & 78.2 & 85.5 & 86.5 & 60.9 \\
SeqVLAD \cite{Mereu_2022_seqvlad}                               & CCT384    &24576     & 2.02  & 17.2 & 6.4 &  \underline{81.7} &  \underline{89.4} & \underline{92.8} & 78.7 \\
SeqVLAD + PCA \cite{Mereu_2022_seqvlad}                         & CCT384    & 4096     & 2.02  & 17.2 & 1.1 &81.4 & 89.2 & \textbf{93.3} & 76.8 \\
TimeSformer \cite{Bertasius_2021_timesformer}                           & -         &  768 & 2.34 & 13.5 & 0.2 & 73.8 & 81.5 & 74.9 & 46.8 \\
\textbf{JIST (Ours)}            & ResNet-18 &  512     & 2.04 & 11.1 & 0.1 & \textbf{88.9} & \textbf{90.6} & 91.5 & \textbf{79.0} \\
\bottomrule
\end{tabular}}
\end{center}
\end{table*}

\myparagraph{Datasets}
To assess the soundness of the JIST multi-task training framework, we use the following datasets:

\noindent
\textbullet\; \emph{Mapillary Street-Level Sequences} (MSLS) \cite{Warburg_2020_msls}, is built from various cities around world, split in non-overlapping training, validation and test sets, and consisting of 393k query sequences and 733k for the database (if we consider 5-frames sequences). As the original test set labels are not released by the authors, we follow the splits defined in \cite{Mereu_2022_seqvlad}. 
  \begin{itemize}[noitemsep,topsep=1pt]
    \item \textit{Test set}: Copenhagen, San Francisco
    \item \textit{Val set}: Amsterdam, Manila
    \item \textit{Train set}: all remaining cities
  \end{itemize}
Unless otherwise specified, experiments (train/val/test) are performed on MSLS.

\noindent
\textbullet\; 
\emph{MSLS Melbourne} is the subset of MSLS from the city of Melbourne, and it is commonly used \cite{Garg_2021_seqNet, Garg_2021_SeqMatchNetCL, Mereu_2022_seqvlad} to understand the effect of training only on a single city as opposed to the entire MSLS train set. When the model is trained on Melbourne, the validation and testing are performed on the standard MSLS val and test sets.

\noindent
\textbullet\;
\emph{San Francisco eXtra Large} (SF-XL) \cite{Berton_2022_cosPlace} is a large-scale (41M images) im2im dataset covering the whole city of San Francisco, and it is used as a training set for the CosPlace component of the loss. Note that CosPlace requires camera heading labels, meaning that most other datasets (MSLS included) can not be used for training CosPlace.

\noindent
\textbullet\;
\emph{Oxford RobotCar} \cite{Maddern_2017_robotCar} is a small dataset containing roughly 4k queries and database sequences in each split. It contains multiple traversals of the same path around the city of Oxford. Laps are recorded in different times of the day, year, as well as changing weather conditions, targeting robustness to domain shifts. In the literature there is little consistency upon which splits to adopt \cite{Garg_2020_deltaDescr, Garg_2021_seqNet, Garg_2021_SeqMatchNetCL},  thus as with MSLS we follow the proposed one in \cite{Mereu_2022_seqvlad}. 
    \begin{compactitem}
      	\item \textit{RobotCar Test set}:
    	queries:  2014-12-16-18-44-24 (winter night); 
        database:  2014-11-18-13-20-12 (fall day).
    	\item \textit{RobotCar Validation set}:
    	queries:  2015-02-03-08-45-10 (winter day, snow);
        database:  2015-11-13-10-28-08 (fall day, overcast).
        \item \textit{RobotCar Train set}: 
        queries:    2014-12-17-18-18-43 (winter night, rain);
    	database:  2014-12-16-09-14-09 (winter day, sun).
    \end{compactitem}

\myparagraph{Training}
For training, we set $\lambda_{im2im}=100$ and $\lambda_{seq2seq}=10.000$.
The learning rate is set to 0.00001 and we use Adam \cite{Kingma_2014_adam} as optimizer.
We train our model for a fixed number of iterations, namely 12.5k.
To speed up convergence and reduce carbon footprint of our trainings, we initialize the backbone with the open-source pretrained weights from CosPlace.
Regarding our architecture, 
we use a ResNet-18 \cite{He_2016_resnet} backbone which has an output dimensionality $D=512$.
We keep the same dimension after the linear projection $D'=512$, except for experiments in \cref{tab:out_dim} where we show that our method works well also with smaller descriptors.
The parameter $p$ of SeqGeM is initialized to 3.

\myparagraph{Evaluation}
We use a standard kNN to find the predictions for each query.
As metric, we use the Recall@N, defined as the number of queries that have at least one correct positives within the first N predictions. A prediction is deemed correct if at least one of its frames is less than 25 meters away from at least one the query's frames, following \cite{Warburg_2020_msls}'s definition of seq2seq.
Unless otherwise specified we use a sequence length of 5 following previous work \cite{Mereu_2022_seqvlad}, although in \cref{fig:seq_len} we show that SeqGeM is able to produce robust descriptors even with different sequence lengths.
Given that in VPR it is logical to either train and test on different (non-overlapping) geographical areas \cite{Arandjelovic_2018_netvlad}, or to consider the train and test sets to be geographically overlapping \cite{Berton_2022_cosPlace}, we compute results for both cases:
results on MSLS use geographically disjointed sets, whereas results on RobotCar use the same area for training and testing.

\myparagraph{Methods}
We report results from a large number of methods on the task of seq2seq VPR.
Wherever available, we made use of the authors official code for our comparisons.
For methods based on the traditional \textit{sequence matching}, we compare against three popular implementations: SeqSLAM \cite{Milford_2012_seqslam}, HVPR \cite{Garg_2021_seqNet}, and SeqMatchNet~\cite{Garg_2021_SeqMatchNetCL}.
We also compare to existing methods based on \emph{sequence descriptors}.
Starting from the work of ~\cite{Warburg_2020_msls}, we test standard concatenation (CAT) of popular im2im descriptors NetVLAD \cite{Arandjelovic_2018_netvlad} and GeM \cite{Radenovic_2019_gem} using different backbones.
We also compute results with Delta Decriptors~\cite{Garg_2020_deltaDescr}, a non-learnt pooling in this category.
We compare against Fully-Connected layers on top of flattened frame descriptors \cite{Facil_2019_multiViewNet}, varying the feature extractor.
Additionally, we test the learnable pooling of SeqNet \cite{Garg_2021_seqNet} and the previous SOTA represented by SeqVLAD \cite{Mereu_2022_seqvlad}. 
Finally, we test a method that processes all frames as a single entity from the first layers,
namely the TimeSformer \cite{Bertasius_2021_timesformer}.

For methods that produce huge descriptors, mostly due to NetVLAD applied on each frame of the sequences, we followed \cite{Mereu_2022_seqvlad} and applied PCA for dimensionality reduction.
It is noteworthy in this sense that our proposed pipeline naturally outputs compact descriptors (512-D) freeing ourselves from the extra cost of applying PCA, while also achieving higher results despite the lower dimensionality.

A few methods (HVPR, SeqMatchNet and SeqNet) could not be trained on the whole MSLS due to large memory requirements of their implementation (more than 256 GB of RAM), hence why some results are missing.
Finally, we clarify that official code for Delta Descriptors and SeqSLAM do not train frame-level descriptors and rely on pre-trained networks. In the table they are highlighted with {*}.


\subsection{Results and discussion}

\begin{figure}[t!]
    \begin{center}
    \begin{minipage}{.22\textwidth}
        \includegraphics[width=\textwidth]{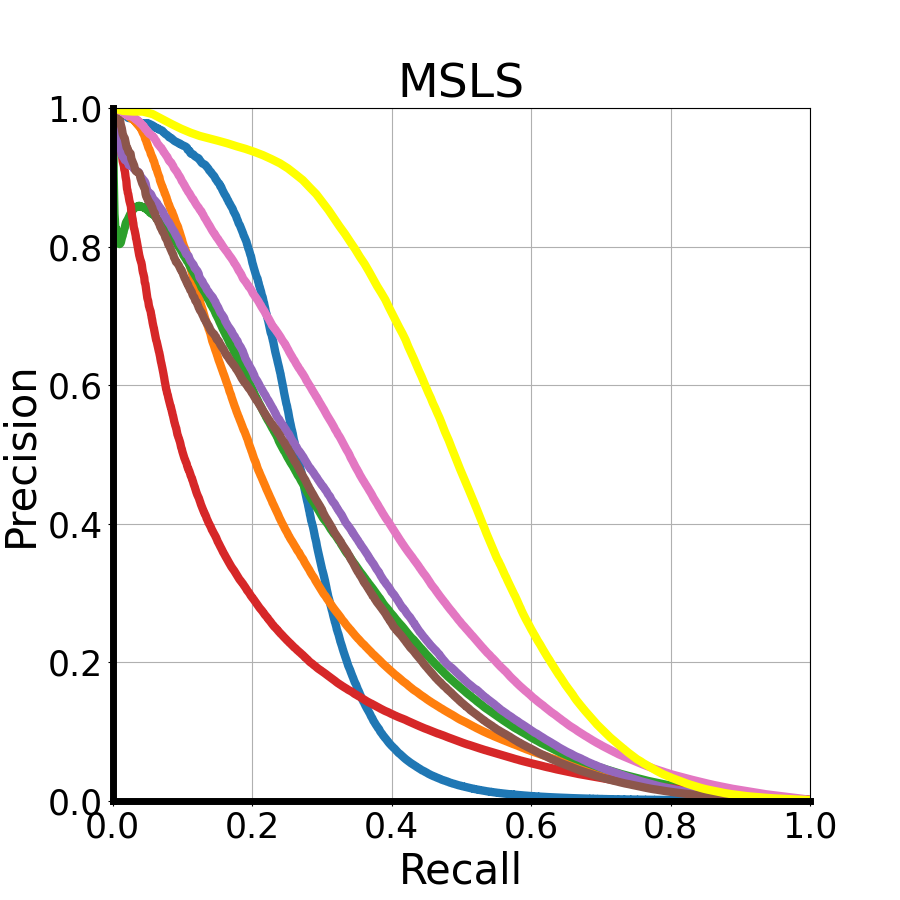}
    \end{minipage}
    \begin{minipage}{.14\textwidth}
        \includegraphics[width=\textwidth]{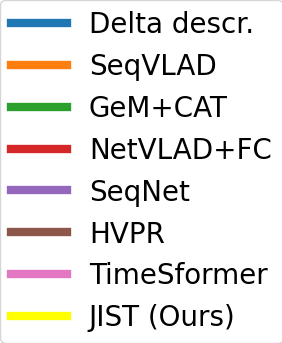}
    \end{minipage}
    \end{center}
\caption{Precision-Recall curves computed on MSLS test set for the most relevant methods. All models are trained with a ResNet-18 backbone except TimeSformer, which uses a custom backbone.}
\label{fig:pr_curve}
\end{figure}

To empirically assess the effectiveness of our proposed models against previous literature, we report a wide set of experiments in \cref{tab:main_tab}, and precision-recall curves for the most relevant methods in \cref{fig:pr_curve}.

We summarize the findings from experiments as follows:
\begin{itemize}[noitemsep,topsep=1pt]
  \item JIST achieves excellent results with small-dimensional descriptors, even when trained on fewer sequential data (\emph{i.e.} training on Melbourne);
  \item SeqVLAD achieves overall good results, but its recalls are poor when trained on fewer data;
  \item Despite its strong results, JIST is extremely fast and uses a simple model for inference;
  \item Extraction time depends mostly on the backbone, and only slightly depend on the aggregation layer (\emph{e.g.} CAT, SeqGeM, FC);
  \item On all considered testing datasets, extraction is the bottleneck, although for a bigger dataset matching would be slower, as its speed linearly depends on dataset size;
  \item We empirically verified that matching time is linearly correlated to descriptors dimension for sequence descriptors (\emph{i.e.} pure retrieval) methods;
\end{itemize}

\myparagraph{Computational cost}
Besides being fast to train (less than 10 hours on a single GPU), JIST provides very efficient inference, due to small descriptors and lightweight architecture.
Specifically, we rely on a ResNet-18, which has only 11M parameters, leading to fast features extraction time.

Matching time is also small (8 times smaller than previous SOTA), due to SeqGeM's compact output:
in fact the matching time (\emph{i.e.} time it takes to find the matching descriptors to the query's through a kNN) depends only on the descriptors' dimension and the size of the database.
Note that, as we scale to larger datasets (with more sequences in the database), the bottleneck of a VPR system at inference shifts from the extraction to matching, making compact descriptors and fast matching an important characteristic for large-scale deployment~\cite{Berton_2022_benchmark}.

\begin{table}[t!]
\caption{Ablation on the two components of the multi-task loss, on MSLS. Best results in bold.}
\label{tab:losses_ablation}
\begin{adjustbox}{width=.99\columnwidth,center}
\begin{tabular}{c|ccccc|ccccc}
\toprule
$\lambda_{seq2seq}$ & 0 & 100 & 1000 & 10k & 100k & 10k & 10k & 10k & 10k & 10k \\
\midrule
$\lambda_{im2im}$ & 100 & 100 & 100 & 100 & 100 & 0 & 1 & 10 & 100 & 1000 \\
\midrule
R@1 & 87.6 & 88.4 & 89.4 & \textbf{90.6} & 90.4 & 89.9 & 90.2 & 90.2 & \textbf{90.6} & 89.9 \\
\bottomrule
\end{tabular}
\end{adjustbox}
\end{table}
\myparagraph{Ablation on the loss}
In this paragraph we aim at understanding how each component of the loss affects results, to justify their use in training.
In \cref{tab:losses_ablation} we report results computed with different weights for $\lambda_{seq2seq}$ and $\lambda_{im2im}$,
with a ResNet-18 and our proposed SeqGeM layer.
We find that when any of the two has a null effect on the back-propagated gradients, the results are evidently lower, proving that both learning objectives are beneficial to the task.
Note that using $\lambda_{seq2seq}=0$ means that only the im2im loss is used (therefore SeqGeM is not trained, but simply initialized to 3).
The best results are shown with values of $\lambda_{seq2seq}=10.000$ and $\lambda_{im2im}=100$.
Finally, we note that the $\mathcal{L}_{seq2seq}$ has a stronger effect than the $\lambda_{im2im}$, as not using the $\mathcal{L}_{seq2seq}$ leads to a 3\% points in reduction with respect to the best model. This effect proves the fact that while it is possible to learn to extract salient features for localization using only single images, a loss that instructs the model how to aggregate temporal information is necessary.

\begin{table}[!h]
\caption{Effect of descriptor dimensionality: JIST vs. SeqVLAD. Best result overall in bold, best per descriptor dimension is underlined.}
\label{tab:out_dim}
\begin{adjustbox}{width=.7\columnwidth,center}
\begin{tabular}{c|ccccc}
\toprule
Descriptor Dim. & 64 & 128 & 256 & 512 & 4096 \\
\midrule
\textbf{JIST(ours)}            & \underline{83.6} & \underline{87.9} & \underline{89.4} & \textbf{90.6} & -\\
SeqVLAD + PCA & 83.4& 85.6& 86.7& 87.7 & 89.2\\
\bottomrule
\end{tabular}
\end{adjustbox}
\end{table}

\myparagraph{Effect of descriptor dimensionality}
Due to its importance for seq2seq VPR, and for retrieval in general, we perform an ablation on the dimensionality of descriptors.
We find that JIST allows trained models to perform astonishingly well even at very low dimensions, reaching an impressive 87.9\% with 128 dimensional descriptors, which is only 1.5\% lower than previous state of the art while being 192 times smaller.
This practically means that the 128D SeqGeM configuration requires only 512 bytes to store each sequence, allowing to store entire cities within a single embedded device.
More considerations on this topic of real-world applicability are in \cref{sec:realworld_deployment}.

\begin{figure}
  \centering
  \includegraphics[width=0.9\columnwidth]{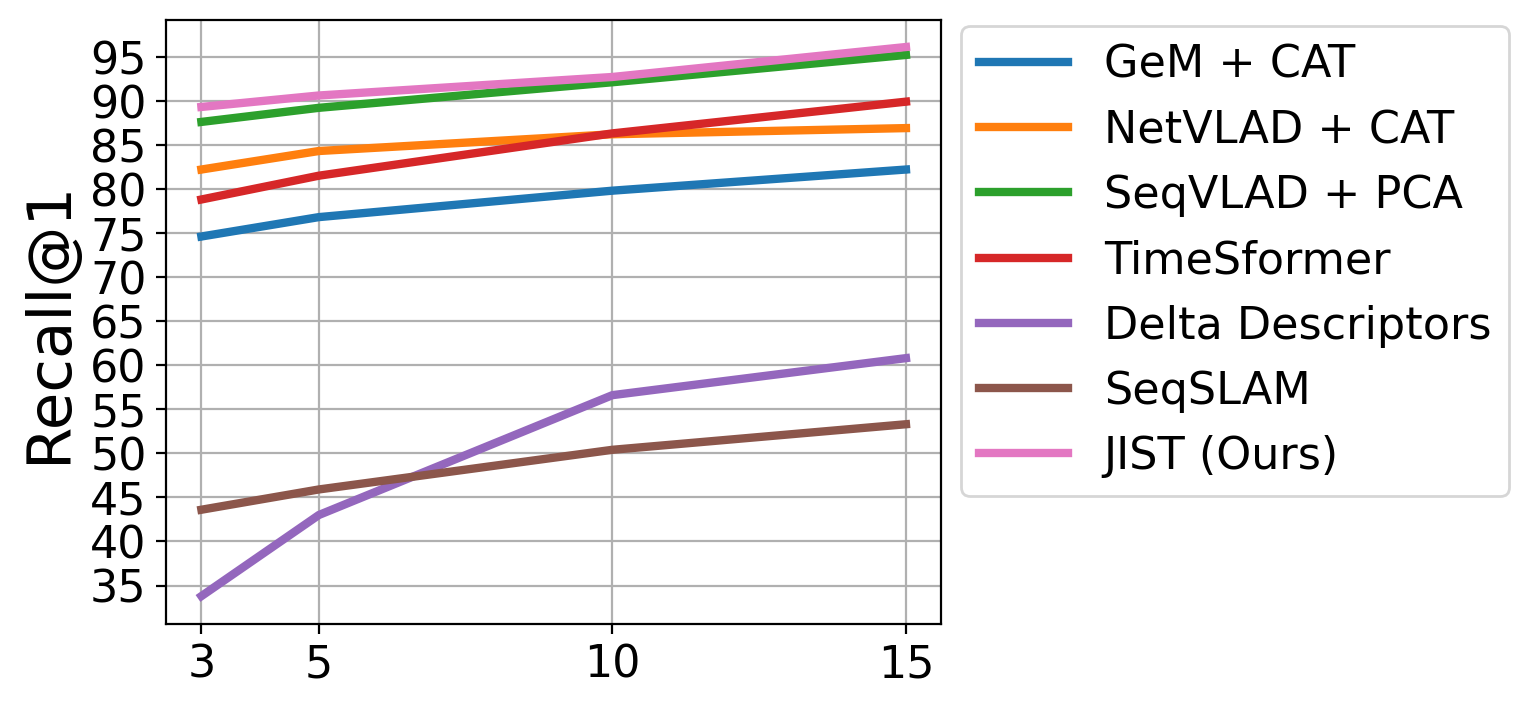}
  \caption{The plot shows how different methods react to changes in the dimension of test-time sequence length (\emph{i.e.} number of frames). All methods are trained with fixed sequence length of 5.}
  \label{fig:seq_len}
\end{figure}

\myparagraph{Effect of sequence length}
In \cref{fig:seq_len} we investigate the effect of changing the number of frames within sequences (sequence length) at test time, without re-training the model, noting that flexibility on processing sequences of arbitrary lengths (regardless of the length at training time) is a desirable property in practical applications.
Firstly, we note that only a small number of methods can be applied to this scenario, i.e. those based on CAT, SeqVLAD, SeqGeM and TimeSformer:
others, like those based on a fully connected layer, would need to be trained again from scratch, as their number of parameters depends on the sequence length.

Clearly, all methods benefit from longer sequences: with more frames, descriptors become more informative, limiting perceptual aliasing.
Models trained with JIST outperform all competitors, especially with very short sequences: this is expected behaviour, as the image loss allows to extract informative features even from a single frame.

\begin{table}[t!]
\caption{Robustness to the inversion of the frames, as R@1. Best (lowest) differences when inverting frames in bold.}
\label{tab:reverse_frames}
\begin{adjustbox}{width=.9\columnwidth,center}
\begin{tabular}{cccccc}
\toprule
Method & Backbone & Dim. & Forw. & Back. & Diff \\
\midrule
SeqSLAM* \cite{Milford_2012_seqslam}                  & VGG-16    & 4096 & 45.9 & 22.9 & -50 \% \\
Delta D.*\cite{Garg_2020_deltaDescr}     & VGG-16    & 4096 & 43.0 & 11.7 & -73 \% \\
HVPR \cite{Garg_2021_seqNet}             & VGG-16    & 4096 & 51.0 & 28.5 & -44 \% \\
SeqMatchNet \cite{Garg_2021_SeqMatchNetCL} & VGG-16    & 4096 & 44.8 & 24.2 & -46 \% \\
GeM + CAT \cite{Warburg_2020_msls}       & ResNet-18 & 1280 & 76.8 & 67.8 & -12 \% \\
NetVLAD + CAT + PCA             & ResNet-18 & 4096 & 83.7 & 79.9 & -5 \%  \\
SeqNet \cite{Garg_2021_seqNet}           & VGG-16    & 4096 & 50.1 & 42.0 & -16 \% \\
NetVLAD + FC                    & ResNet-18 & 4096 & 68.5 & 65.1 & -5 \% \\
SeqVLAD + PCA                   & ResNet-18 & 4096 & 85.5 & 85.2 & -0.4 \% \\
SeqVLAD + PCA                   & CCT384    & 4096 & 89.2 & 89.2 & \textbf{0.0} \% \\
TimeSformer                     & -         & 768 & 81.5 & 81.5 & \textbf{0.0} \%\\
\textbf{JIST (Ours)}     & ResNet-18 & 512  & 90.6 & 90.6 & \textbf{0.0} \% \\
\bottomrule
\end{tabular}
\end{adjustbox}
\end{table}

\myparagraph{Effect of reversing frames}
Robustness to frame ordering is a desirable property in some realistic use-cases, because it allows to reduce the number of sequences stored in the database.
Following \cite{Facil_2019_multiViewNet, Garg_2021_seqNet, Mereu_2022_seqvlad}
we assess each model's robustness to reversing the frame ordering for queries sequences, while keeping the database untouched, and report results in \cref{tab:reverse_frames}.
SeqGeM is inherently robust to frame-ordering, as well as SeqVLAD and TimeSformer which processes the sequence in its entirety. On the other hand, methods based on FC-layers, CAT or sequence matching are the ones that suffer most in this scenario.

\begin{table}[t!]
\caption{\textbf{Comparison of  aggregation layers}.
Recall@1 is computed with same training configuration on MSLS splits.
}
\label{tab:ablation_aggregations}
\begin{adjustbox}{width=.99\columnwidth,center}
\begin{tabular}{c|ccccccccccc}
\toprule
\multirow{2}{*}{\begin{tabular}[c]{@{}c@{}}Aggregation\end{tabular}} &
\multirow{2}{*}{\begin{tabular}[c]{@{}c@{}}Learnable\end{tabular}} &
\multirow{2}{*}{\begin{tabular}[c]{@{}c@{}}Flexible w.r.t.\\length input seq.\end{tabular}} &
\multirow{2}{*}{\begin{tabular}[c]{@{}c@{}}Invariant to\\frame order\end{tabular}} &
\multirow{2}{*}{\begin{tabular}[c]{@{}c@{}}Output\\Dimensionality\end{tabular}} &
\multirow{2}{*}{\begin{tabular}[c]{@{}c@{}}R@1\end{tabular}}
\\ \\
\midrule
Max Pooling & N & Y & Y & 512 & 89.4 \\
Avg Pooling & N & Y & Y & 512 & 89.5 \\
1D-conv     & Y & N & N & 512 & 88.7 \\
SeqVLAD     & Y & Y & Y & 32768 & 89.7 \\
SeqGeM      & Y & Y & Y & 512 & 90.6 \\
\bottomrule
\end{tabular}
\end{adjustbox}
\end{table}

\myparagraph{Ablation on aggregation layer}
Given their importance in aggregating features from multiple frames, in \cref{tab:ablation_aggregations} we report experiments performed with a number of pooling/aggregation layers. This shows a number of desirable properties that are satisfied by SeqGem, as well as showing its superiority of results.
In particular, we note that the SeqGeM aggregator provides the following characteristics: 
(i) learnable, (ii) flexible w.r.t. length of input sequences, (iii) invariant to frame ordering, (iv) lightweight, besides producing compact output and having few parameters.

Note that the results from \cref{tab:ablation_aggregations} are performed within the JIST framework/pipeline, making these aggregations achieve superior recalls w.r.t. most of the baselines from \cref{tab:main_tab}.


\subsection{Considerations for real-world deployment}
\label{sec:realworld_deployment}
As the use of deep models for seq2seq VPR becomes widespread, we investigate the feasibility of deploying such models in the real world. We perform experiments on a Jetson Nano platform. 
Considering the scenario of a large city like San Francisco, with 1600 kilometers of road, it would require roughly 800k sequences to map the whole city.
Using the previous state-of-the-art model, namely CCT384 \cite{Hassani_2021_cct} with SeqVLAD \cite{Mereu_2022_seqvlad}, it needs $\approx 36GB$ (\emph{\#sequences * descriptors dimension * \#bytes}) of memory to store all the descriptors.
More compact representations (commonly compressed with PCA) usually rely on 4096-D features \cite{Mereu_2022_seqvlad}, at the cost of a performance penalty.
With SeqGeM however, we are able to outperform previous state of the art with 512-D descriptors, which needs only $800k * 512 * 4B \approx 0.75GB$, and can be handled by a Jetson Nano.

Given this setting, we analyzed the inference time on a Jetson Nano: we found that extraction time for a sequence takes 276 ms (i.e. with our ResNet-18; does not depend on the size of the database). Matching takes 3.1 seconds with a vanilla kNN (on the whole city of San Francisco).
We note that previous works on im2im VPR found that kNN can be sped up by up to 64 times with negligible loss of recall \cite{Berton_2022_benchmark} when using approximate/efficient versions of it, like Inverted File Index with Product Quantization \cite{Jegou_2011_productQ, Babenko_2014_inv_multiindex}, leading to a potential processing speed of roughly 3 sequences per second ($276 ms + (3100 / 64) ms = 324 ms$), whereas previous SOTA (with descriptors dimension 24576) would process only 0.4 sequences per second. Even with PCA, the throughput would still be limited to 1.4 sequences per second.

\section{Conclusion}

This work proposes a novel training algorithm that efficiently exploits existing data sources to boost performance in sequence-based VPR.
We introduce a trainable temporal aggregation layer designed to being flexible to input length and frame ordering, all while guaranteeing compact descriptors.
Through extensive experimental evaluation we showcase the improvements that JIST achieves over previous SOTA, as well as robustness to different conditions such as changes in frame ordering, sequence length and different datasets. 
We empirically demonstrate that our model is able to not only achieve better results, but also be faster and lighter (in terms of RAM and GPU memory).

\myparagraph{Limitations}
Although  sequence descriptors are a competitive solution to obtain efficiently a coarse global localization estimate even in very large environments, their use is intended when there is the need to search in a large number of sequences (e.g., for loop closure or to bootstrap the localization when lost).
Furthermore, we note that a limitation of the current JIST framework is that the two losses require different format of datasets, where the im2im branch is trained on large-scale single-image datasets whereas the seq2seq branch requires continual sequences.

\myparagraph{Future works}
Possible directions for follow-up works may explore different strategies for extracting knowledge from large pre-trained models (e.g. distillation), generalizing our multi-task framework to other tasks, or using more than two branches to gather knowledge from other data sources.

\myparagraph{Acknowledgements} We acknowledge the CINECA award under the ISCRA initiative, for the availability of high performance computing resources. This work was supported by CINI.




\bibliographystyle{IEEEtran}
\bibliography{bibliography}

\begin{thebibliography}{10}
\providecommand{\url}[1]{#1}
\csname url@rmstyle\endcsname
\providecommand{\newblock}{\relax}
\providecommand{\bibinfo}[2]{#2}
\providecommand\BIBentrySTDinterwordspacing{\spaceskip=0pt\relax}
\providecommand\BIBentryALTinterwordstretchfactor{4}
\providecommand\BIBentryALTinterwordspacing{\spaceskip=\fontdimen2\font plus
\BIBentryALTinterwordstretchfactor\fontdimen3\font minus
  \fontdimen4\font\relax}
\providecommand\BIBforeignlanguage[2]{{%
\expandafter\ifx\csname l@#1\endcsname\relax
\typeout{** WARNING: IEEEtran.bst: No hyphenation pattern has been}%
\typeout{** loaded for the language `#1'. Using the pattern for}%
\typeout{** the default language instead.}%
\else
\language=\csname l@#1\endcsname
\fi
#2}}

\bibitem{Masone_2021_survey}
C.~Masone and B.~Caputo, ``A survey on deep visual place recognition,''
  \emph{IEEE Access}, vol.~9, pp. 19\,516--19\,547, 2021.

\bibitem{Konstantinos-2022_loopclosure}
K.~A. Tsintotas, L.~Bampis, and A.~Gasteratos, ``The revisiting problem in
  simultaneous localization and mapping: A survey on visual loop closure
  detection,'' \emph{IEEE Trans. on Pattern Anal. and Mach. Intell.}, 2022.

\bibitem{Angeli-2008-relocalization}
A.~Angeli, S.~Doncieux, J.-A. Meyer, and D.~Filliat, ``Real-time visual
  loop-closure detection,'' in \emph{IEEE Int. Conf. on Robot. and Autom.},
  2008, pp. 1842--1847.

\bibitem{Pion_2020_benchmark}
N.~Pion, M.~Humenberger, G.~Csurka, Y.~Cabon, and T.~Sattler, ``Benchmarking
  image retrieval for visual localization,'' in \emph{2020 International
  Conference on 3D Vision (3DV)}, 2020, pp. 483--494.

\bibitem{Yudin_2023_indoor_vpr}
D.~Yudin, Y.~Solomentsev, R.~Musaev, A.~Staroverov, and A.~I. Panov,
  ``Hpointloc: Point-based indoor place recognition using synthetic rgb-d
  images,'' in \emph{Neural Information Processing}, 2023, pp. 471--484.

\bibitem{Zeng_2017_underground_vpr}
F.~Zeng, A.~Jacobson, D.~W. Smith, N.~Boswell, T.~Peynot, and M.~Milford,
  ``Enhancing underground visual place recognition with shannon entropy
  saliency,'' in \emph{IEEE Int. Conf. on Robot. and Autom.}, 2017.

\bibitem{Toft_2020_visuallocalization_net}
C.~Toft, W.~Maddern, A.~Torii, L.~Hammarstrand, E.~Stenborg, D.~Safari,
  M.~Okutomi, M.~Pollefeys, J.~Sivic, T.~Pajdla, F.~Kahl, and T.~Sattler,
  ``Long-term visual localization revisited,'' \emph{IEEE Trans. on Pattern
  Anal. and Mach. Intell.}, 2020.

\bibitem{Warburg_2020_msls}
F.~Warburg, S.~Hauberg, M.~Lopez-Antequera, P.~Gargallo, Y.~Kuang, and
  J.~Civera, ``Mapillary street-level sequences: A dataset for lifelong place
  recognition,'' in \emph{Conf. on Comput. Vis. and Pattern Recog.}, 2020, pp.
  2623--2632.

\bibitem{Facil_2019_multiViewNet}
J.~M. F{\'a}cil, D.~Olid, L.~Montesano, and J.~Civera, ``Condition-invariant
  multi-view place recognition,'' \emph{ArXiv}, vol. abs/1902.09516, 2019.

\bibitem{Garg_2020_deltaDescr}
S.~Garg, B.~Harwood, G.~Anand, and M.~Milford, ``Delta descriptors:
  Change-based place representation for robust visual localization,''
  \emph{IEEE Robot. and Autom. Letters}, vol.~5, no.~4, pp. 5120--5127, 2020.

\bibitem{Garg_2021_seqNet}
S.~Garg and M.~Milford, ``{SeqNet}: Learning descriptors for sequence-based
  hierarchical place recognition,'' \emph{IEEE Robot. and Autom. Letters},
  2021.

\bibitem{Mereu_2022_seqvlad}
R.~Mereu, G.~Trivigno, G.~Berton, C.~Masone, and B.~Caputo, ``Learning
  sequential descriptors for sequence-based visual place recognition,''
  \emph{IEEE Robot. and Autom. Letters}, vol.~7, no.~4, pp. 10\,383--10\,390,
  2022.

\bibitem{Berton_2022_cosPlace}
G.~Berton, C.~Masone, and B.~Caputo, ``Rethinking visual geo-localization for
  large-scale applications,'' in \emph{Conf. on Comput. Vis. and Pattern
  Recog.}, 2022, pp. 4868--4878.

\bibitem{Alibey_2023_mixvpr}
A.~Ali-bey, B.~Chaib-draa, and P.~Gigu{\`e}re, ``Mixvpr: Feature mixing for
  visual place recognition,'' in \emph{IEEE Wint. Conf. on Appl. of Comput.
  Vis.}, 2023, pp. 2998--3007.

\bibitem{Berton_2022_benchmark}
G.~Berton, R.~Mereu, G.~Trivigno, C.~Masone, G.~Csurka, T.~Sattler, and
  B.~Caputo, ``Deep visual geo-localization benchmark,'' in \emph{Conf. on
  Comput. Vis. and Pattern Recog.}, 2022, pp. 5386--5397.

\bibitem{Philbin_2007_oxford5k}
J.~Philbin, O.~Chum, M.~Isard, J.~Sivic, and A.~Zisserman, ``Object retrieval
  with large vocabularies and fast spatial matching.'' in \emph{Conf. on
  Comput. Vis. and Pattern Recog.}, 2007, pp. 1--8.

\bibitem{Martinez_2020_Pit30M}
J.~Martinez, S.~Doubov, J.~Fan, l.~A. Bârsan, S.~Wang, G.~Máttyus, and
  R.~Urtasun, ``Pit30m: A benchmark for global localization in the age of
  self-driving cars,'' in \emph{2020 IEEE/RSJ International Conference on
  Intelligent Robots and Systems (IROS)}, 2020, pp. 4477--4484.

\bibitem{Radenovic_2019_gem}
F.~Radenovi{\'c}, G.~Tolias, and O.~Chum, ``{Fine-tuning {CNN} Image Retrieval
  with No Human Annotation},'' \emph{IEEE Trans. on Pattern Anal. and Mach.
  Intell.}, 2018.

\bibitem{Ho_2007_first_seq_search}
K.~L. Ho and P.~Newman, ``Detecting loop closure with scene sequences,''
  \emph{Int. J. Comput. Vis.}, vol.~74, no.~3, pp. 261--286, 2007.

\bibitem{Milford_2012_seqslam}
M.~J. Milford and G.~F. Wyeth, ``{SeqSLAM}: Visual route-based navigation for
  sunny summer days and stormy winter nights,'' in \emph{IEEE Int. Conf. on
  Robot. and Autom.}, 2012, pp. 1643--1649.

\bibitem{Schubert_2021_vpr_hard}
S.~Schubert and P.~Neubert, ``What makes visual place recognition easy or
  hard?'' \emph{ArXiv}, vol. abs/2106.12671, 2021.

\bibitem{Naseer_2018_vl_across_seasons}
T.~{Naseer}, W.~{Burgard}, and C.~{Stachniss}, ``Robust visual localization
  across seasons,'' \emph{IEEE Transactions on Robotics}, vol.~34, no.~2, pp.
  289--302, 2018.

\bibitem{Vysotska-2016_lazymatching}
O.~Vysotska and C.~Stachniss, ``Lazy data association for image sequences
  matching under substantial appearance changes,'' \emph{IEEE Robot. and Autom.
  Letters}, vol.~1, no.~1, pp. 213--220, 2016.

\bibitem{Schubert-2021_graphbased}
S.~Schubert, P.~Neubert, and P.~Protzel, ``Graph-based non-linear least squares
  optimization for visual place recognition in changing environments,''
  \emph{IEEE Robotics and Automation Letters}, vol.~6, no.~2, pp. 811--818,
  2021.

\bibitem{schubert-2021_fast}
S.~Schubert, P.~Neubert, and Protzel, ``Fast and memory efficient graph
  optimization via icm for visual place recognition.'' in \emph{Robotics:
  Science and Systems}, vol.~73, 2021, pp. 1842--1847.

\bibitem{Garg_2021_SeqMatchNetCL}
S.~Garg, M.~Vankadari, and M.~Milford, ``{SeqMatchNet}: Contrastive learning
  with sequence matching for place recognition \& relocalization,'' in
  \emph{CoRL}, ser. Proc. Mach. Learn. Res., vol. 164.\hskip 1em plus 0.5em
  minus 0.4em\relax PMLR, 2022, pp. 429--443.

\bibitem{Neubert-2021_hyperd}
P.~Neubert and S.~Schubert, ``Hyperdimensional computing as a framework for
  systematic aggregation of image descriptors,'' in \emph{2021 Conf. on Comput.
  Vis. and Pattern Recog.}, 2021, pp. 16\,933--16\,942.

\bibitem{zhang_2023_spatiotemp}
F.~Zhang, J.~Zhao, Y.~Cai, G.~Tian, W.~Mu, and C.~Ye, ``Learning sequence
  descriptor based on spatiotemporal attention for visual place recognition,''
  \emph{arXiv preprint arXiv:2305.11467}, July 2023.

\bibitem{Arandjelovic_2018_netvlad}
R.~{Arandjelović}, P.~Gronat, A.~Torii, T.~Pajdla, and J.~Sivic, ``{NetVLAD}:
  {CNN} architecture for weakly supervised place recognition,'' \emph{IEEE
  Trans. on Pattern Anal. and Mach. Intell.}, vol.~40, no.~6, pp. 1437--1451,
  2018.

\bibitem{Alibey_2022_gsvcities}
A.~Ali-bey, B.~Chaib-draa, and P.~Gigu{\`e}re, ``Gsv-cities: Toward appropriate
  supervised visual place recognition,'' \emph{Neurocomputing}, 2022.

\bibitem{Leyvavallina_2021_gcl}
M.~Leyva-Vallina, N.~Strisciuglio, and N.~Petkov, ``Data-efficient large scale
  place recognition with graded similarity supervision,'' \emph{Conf. on
  Comput. Vis. and Pattern Recog.}, pp. 23\,487--23\,496, 2023.

\bibitem{Wang_2018_cosFace}
H.~Wang, Y.~Wang, Z.~Zhou, X.~Ji, D.~Gong, J.~Zhou, Z.~Li, and W.~Liu,
  ``Cosface: Large margin cosine loss for deep face recognition,'' in
  \emph{Conf. on Comput. Vis. and Pattern Recog.}, 2018, pp. 5265--5274.

\bibitem{Hassani_2021_cct}
A.~Hassani, S.~Walton, N.~Shah, A.~Abuduweili, J.~Li, and H.~Shi, ``Escaping
  the big data paradigm with compact transformers,'' \emph{CoRR}, vol.
  abs/2104.05704, 2021.

\bibitem{Bertasius_2021_timesformer}
G.~Bertasius, H.~Wang, and L.~Torresani, ``Is space-time attention all you need
  for video understanding?'' in \emph{Int. Conf. Mach. Learn.}, 2021, pp.
  813--824.

\bibitem{Maddern_2017_robotCar}
W.~Maddern, G.~Pascoe, C.~Linegar, and P.~Newman, ``{1 Year, 1000km: The Oxford
  RobotCar Dataset},'' \emph{The Int. J. of Robot. Research}, 2017.

\bibitem{Kingma_2014_adam}
D.~Kingma and J.~Ba, ``Adam: A method for stochastic optimization,'' \emph{Int.
  Conf. Learn. Represent.}, 2014.

\bibitem{He_2016_resnet}
K.~{He}, X.~{Zhang}, S.~{Ren}, and J.~{Sun}, ``Deep residual learning for image
  recognition,'' in \emph{Conf. on Comput. Vis. and Pattern Recog.}, 2016, pp.
  770--778.

\bibitem{Jegou_2011_productQ}
H.~Jégou, M.~Douze, and C.~Schmid, ``Product quantization for nearest neighbor
  search.'' \emph{IEEE Trans. on Pattern Anal. and Mach. Intell.}, vol.~33,
  no.~1, pp. 117--128, 2011.

\bibitem{Babenko_2014_inv_multiindex}
A.~Babenko and V.~S. Lempitsky, ``The inverted multi-index.'' in \emph{Conf. on
  Comput. Vis. and Pattern Recog.}\hskip 1em plus 0.5em minus 0.4em\relax IEEE
  Computer Society, 2012, pp. 3069--3076.

\end{thebibliography}

\end{document}